%% file: acl_latex.tex
\pdfoutput=1

\documentclass[11pt]{article}

\usepackage[preprint]{acl}

\usepackage{times}
\usepackage{latexsym}

\usepackage[T1]{fontenc}

\usepackage[utf8]{inputenc}

\usepackage{microtype}

\usepackage{inconsolata}

\usepackage{xcolor}
\usepackage{graphicx}
\usepackage{booktabs}
\usepackage{amsmath}
\usepackage{makecell}
\usepackage{colortbl}
\usepackage{array}
\usepackage{longtable}

\title{Can Large Language Models Generate High-quality Patent Claims?}

\author{
Lekang Jiang$^{\dagger}$, 
Caiqi Zhang$^{\dagger}$, 
Pascal A Scherz$^{\diamond}$,
Stephan Goetz$^{\dagger}$ \\
$^{\dagger}$University of Cambridge, $^{\diamond}$PSPB Patent Law\\ 
\texttt{\{lj408, cz391, smg84\}@cam.ac.uk, post@pspb.eu}
}

\begin{document}
\maketitle
\begin{abstract}

Large language models (LLMs) have shown exceptional performance across various text generation tasks but remain under-explored in the patent domain, which offers highly structured and precise language. This paper constructs a dataset to investigate the performance of current LLMs in patent claim generation. Our results demonstrate that generating claims based on patent descriptions outperforms previous research relying on abstracts. 
Interestingly, current patent-specific LLMs perform much worse than state-of-the-art general LLMs, highlighting the necessity for future research on in-domain LLMs.
We also find that LLMs can produce high-quality first independent claims, but their performances markedly decrease for subsequent dependent claims.
Moreover, fine-tuning can enhance the completeness of inventions' features, conceptual clarity, and feature linkage. 
Among the tested LLMs, GPT-4 demonstrates the best performance in comprehensive human evaluations by patent experts, with better feature coverage, conceptual clarity, and technical coherence. 
Despite these capabilities, comprehensive revision and modification are still necessary to pass rigorous patent scrutiny and ensure legal robustness.\footnote{\url{https://github.com/scylj1/LLM4DPCG}}

\end{abstract}

\section{Introduction}

Large language models (LLMs) have demonstrated remarkable capabilities across a broad spectrum of general tasks, ranging from natural language understanding to complex problem-solving \citep{zhao2023survey,min2023recent}. However, the application of LLMs in specialized domains remains under-explored, such as the patent literature \citep{jiang2024artificial}. Patents, which are legal documents detailing and protecting inventions to promote technical innovations \citep{mossoff2000rethinking}, present unique challenges due to their complex and technical nature. Meanwhile, their language features a precision level that is unheard of in other text corpora. The potential for LLMs to revolutionize this field is significant, offering possibilities to enhance technical knowledge extraction, patent analysis, and generation \citep{jiang2024artificial}. 

\begin{figure}[t]
  \includegraphics[width=0.99\linewidth]{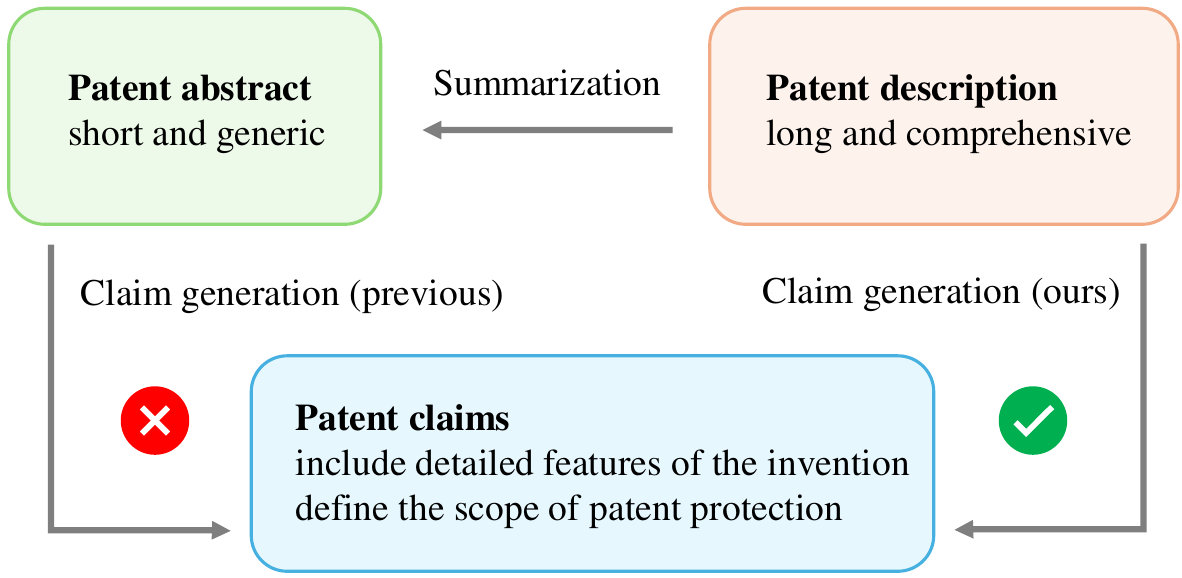}
  \caption{Overview of patent claim generation. Previous abstract-based claim generation is problematic as abstracts lack the detailed technical and legal specifics required for claims. Instead, patent descriptions include all elements or features of the claims, providing a more suitable basis for claim generation.  }
  \label{fig:introduction}
\end{figure}

\input{tabs/comparison_with_summarization}

Patent claims are the critical part of a patent application that defines the boundaries of patent protection. We introduce the background of patent descriptions, abstracts, and claims in Appendix~\ref{examplepatent}. Drafting patent claims necessitates the expertise of qualified patent agents or attorneys. It requires a profound understanding of the invention's technical details and a familiarity with pertinent patent laws and language conventions \citep{faber1990landis}. Meticulous and precise patent claims are essential for securing robust patent protection. However, drafting and revising patent applications are both time-intensive and financially demanding. Each round of revising can generate costs up to thousands of dollars \citep{CisloAndThomas2023}, presenting considerable challenges particularly for small enterprises. Therefore, automation through appropriate assistive language models could significantly enhance quality and efficiency. 

As shown in Figure \ref{fig:introduction}, previous studies have mainly focused on generating claims based on patent abstracts for simplicity \citep{lee2020controlling}. However, the performance of these methods is not satisfying and the task appears ill-posed as abstracts do not fully disclose all details of an invention. In contrast, patent descriptions provide detailed explanations of the invention and its operation. It must, by law, fully disclose the details that the claims summarize and condense into the gist \citep{epo2020}. Therefore, this paper investigates description-based claim generation and compares it with previous solely abstract-based studies. 

Patent claim generation has similar input and output formats as text summarization  \citep{pu2023summarization}. Both require the understanding of long and complicated documents and the generation of shorter target information. However, patent claim generation is substantially more complex and difficult. We list the similarities and differences between claim generation and summarization in 
Table~\ref{table:comparison}. The key differences include: (1) From the content perspective, patent claims include detailed features of the invention, while the abstract only provides a brief summary. Hence, the model needs to understand more nuanced contexts and extract more detailed information for claim generation. (2) Claims have strict writing standards and require high-level precision of wording. The requirement for clarity and precision complicates the task compared to generating less formal and more narrative abstracts. (3) The claim set is a structured list of independent and dependent claims. The logical linkage between different invention's features necessitates the potential reasoning capabilities. 

\input{tabs/patent_datasets}

Since drafting patent claims by qualified patent agents or attorneys is both time-consuming and financially burdensome, we aim to explore whether this process can be realized by LLMs. We design three research questions (RQs) to explore whether LLMs can generate high-quality patent claims.

\noindent \textbf{RQ1:} Does description-based claim generation outperform abstract-based generation?
In principle, patent claims can be generated from any part of a patent document. Previous studies have focused on using patent abstracts for claim generation, despite their brevity and limited legal significance. Since the description is the legal centerpiece, we proposed claim generation based on the description rather than other parts and compared it with previous abstract-based methods.\footnote{It is worth noting that combining abstract and description is neither legally nor technically judicious. Legally: the abstract does not contribute extra information to the description and if it did, it must not be used in the claims (37 US CFR 1.72 and 1.438/PCT Rule 8); the claims have to be solely rooted in the description. Technically: the use of the abstract is rather a wish driven by model limitations; as soon as a model can manage the description, the abstract entirely loses its value as it cannot contain any useful content beyond the description legally anyway; if a model has additional bandwidth for input, there may rather be the question if it could be used for adding external information, e.g., close prior art or similar. Future research should rather study that. Therefore, inputting both abstract and description may be unnecessary and even complicate the task.}

\noindent \textbf{RQ2:} Does current patent-specific LLM outperform general LLMs on claim generation?
It is shown that small-scale domain-specific LLMs (e.g., 7B) can outperform strong general LLMs (e.g., GPT-4) on specific tasks, such as in math reasoning \citep{shao2024deepseekmath} and fact-checking \citep{zhang-etal-2024-need}. The legal-specific LLM SaulLM-7B is also reported to outperform Mistral-7B and Llama-2-13B on LegalBench-Instruct and Legal-MMLU benchmarks \citep{colombo2024saullm}. Therefore, we included legal and patent-specific LLMs to investigate whether domain-specific LLMs can perform better on this task.

\noindent \textbf{RQ3:} How do LLMs perform on patent claims’ completeness, clarity, consistency, and logical linkage?
To make comprehensive evaluations, we have tested with different model architectures, different sizes, and models in specific domains. We conduct human evaluations that adhere strictly to established examination criteria to assess LLMs' capabilities and limitations more accurately.

The main contributions are as follows:

\noindent 1. We construct the first dataset for description-based claim generation. We demonstrate theoretically and empirically that description-based claim generation outperforms previous abstract-based generation, particularly increasing the invention's feature completeness. 

\noindent 2. We show that current patent-specific LLMs perform much worse than state-of-the-art general LLMs, indicating the need for future research in the domain.  We also find LLMs can generate high-quality first independent claims, but underperform when producing subsequent dependent claims. Fine-tuning can improve feature completeness, conceptual clarity, and feature linkage, while multi-task fine-tuning decreases conceptual clarity.

\noindent 3. GPT-4 demonstrates better feature coverage, conceptual clarity, and technical coherence. However, thorough revision and adjustment are still necessary to pass legal scrutiny and ensure claims' legal and technical robustness.

\section{Related Work}

\citet{jiang2024artificial} summarized four types of text generation tasks in the patent field, including summarization, translation, simplification, and patent writing. We list open-sourced datasets aiming at patent text generation in Table~\ref{table:datasets}. Patent summarization \citep{sharma-etal-2019-bigpatent} and translation \citep{pouliquen2015full} have been extensively studied. Notably, patent translation systems are provided by various patent offices, including the European Patent Office's Patent Translate.\footnote{ \url{https://www.epo.org/en/searching-for-patents/helpful-resources/patent-translate}.}

Despite the potential of LLMs, current research in patent writing is still nascent and generally falls short of expectations. Previous works have focused on claim generation. A pioneering study by \citet{lee2020patentgenerate} investigated the use of GPT-2 \citep{radford2019language} for generating patent claims, finding that minimal training was sufficient for producing patent-like texts, though without evaluating their quality.  Subsequently, \citet{lee2020controlling} broadened this research by training GPT-2 to transform one component of a patent application into another, such as converting abstracts into claims. As abstracts are normally generic and imprecise, generating claims from abstracts may not be a well-conditioned task. A concurrent work investigated whether LLMs can revise patent claims to improve quality \citep{jiang-etal-2025-patent}. A follow-up study further explored claim generation on European patents \citep{epd2025}.

Apart from claim generation, \citet{christofidellis2022pgt} introduced a prompt-based generative transformer with GPT-2 as a foundational model. They incorporated multi-task learning (MTL) \citep{maurer2016benefit} to facilitate diverse patent-related tasks including part-of-patent generation, text infilling, and evaluating patent coherence. Additionally, \citet{aubakirova2023patfig} developed the first extensive dataset for patent figure-caption generation, aimed at facilitating the creation of figure captions within patents.

\section{Experiment Setup}

\subsection{Dataset}

\input{tabs/dataset_statistics}

In this research, we construct the first dataset for description-based claim generation. The Harvard USPTO Patent Dataset (HUPD) \cite{suzgun2024harvard} is a recently collected large-scale multi-purpose patent dataset, including more than 4.5 million patent documents with 34 data fields (patent description, abstract, and claims are included).\footnote{The USPTO stands for the United States Patent and Trademark Office, granting U.S.\ patents written in English.} We propose our dataset, HUPD-DCG (Description-based Claim Generation), by filtering a portion of target documents based on this large dataset. 

Firstly, we selected all the patent documents filed in 2017. We eliminated any pending or rejected documents and only kept granted documents to formulate a high-quality dataset for claim generation. Considering the context length of some LLMs, such as Llama-3, we opted for documents with a description length smaller than 8,000 tokens in our experiments. In practical settings, models are developed by training on existing patent documents and subsequently employed for new applications. To simulate realistic scenarios, we ordered the documents by date and used the last 1,311 documents for testing, which is about 14\% of the whole dataset.

Table~\ref{table:dataset_statistics} shows detailed statistics of our dataset, indicating that train and test sets have similar data distributions. The abstract is shorter with a word coverage of 0.08 and a compression ratio of 48, while the claims are more detailed with higher coverage and less compression ratio. 

\subsection{Models}

We select the recent Llama-3-8B\footnote{\url{https://llama.meta.com/llama3/}} and Mistral-7B \citep{jiang2023mistral} as the base models because they have shown outstanding capabilities and represent different model structures. To investigate the size effect, we opt for Llama-3-70B, Mixtral-8$\times$7B \citep{jiang2024mixtral}, Mixtral-8$\times$22B, and state-of-the-art GPT-4 \citep{achiam2023gpt} model. In addition, we conduct experiments on domain-specific LLMs, including the largest open-source patent LLM PatentGPT-J-6B \citep{lee2023evaluating}, and the recent legal-specific LLM SaulLM-7B \citep{colombo2024saullm}. Furthermore, we fine-tune the Llama-3-8B model based on our train set. Llama-3-FT is the version of normal single-task fine-tuning, while Llama-3-MFT is trained on two tasks simultaneously using multi-task fine-tuning (MFT), including description-based claim generation and summarization (abstract generation). We introduce details of each model in Appendix \ref{modeldetails} and experimental details in Appendix \ref{experimentsdetail}. 

\begin{figure*}[t]
  \includegraphics[width=0.99\linewidth]{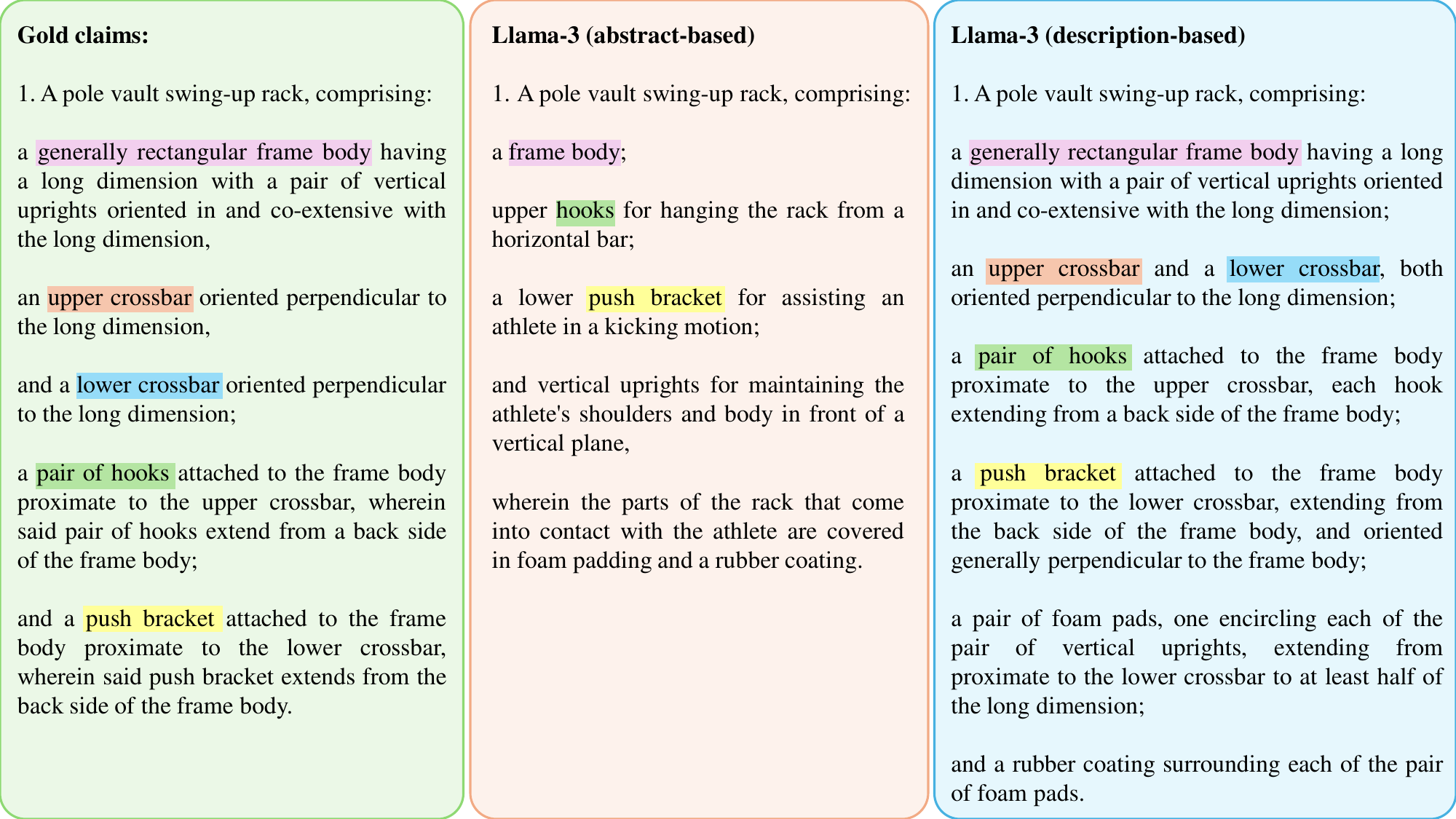}
  \caption{The first claim of patent application US~2018/0064980~A1 (Gold claim, claim generated based on abstract, and claim generated based on description). The key features included in the gold claims are marked in colors. The description-based method generates more precise features comprised in gold claims. }
  \label{fig:abs_claims}
\end{figure*}

\subsection{Evaluation Metrics}

\noindent \textbf{Automated evaluation } We use standard evaluation metrics for text generation, including BLEU \citep{papineni2002bleu}, ROUGE-1 (R-1), ROUGE-L (R-L) \citep{lin2004rouge}, and BERTScore \citep{zhang2019bertscore}. BLEU, R-1, and R-L are surface-level text comparisons, which basically measure the overlap of words or sequences between the generated text and the reference texts, while BERTScore quantifies the semantic similarity. 
 
\noindent \textbf{Human evaluation }
Although automated metrics provide standardized and reproducible ways to compare model performance based on lexical overlaps or semantic relationships, we suspect them to struggle. We supplement our approach with human evaluations conducted by patent professionals. These evaluations adhere strictly to established examination criteria, enhancing the accuracy of our results. Readers should primarily focus on human evaluation results, whereas automated metrics may provide valuable insights at times.

Human evaluation for patent claims needs substantial patent expertise. However, involving patent experts to evaluate large numbers of claim sets is prohibitively costly and labor-intensive.\footnote{Hourly rates per attorney are about \$350 to \$750.} Therefore, we specifically conduct the human evaluation on eighty patent claims of multiple versions generated by LLMs.\footnote{Since the outputs generated by domain-specific LLMs were mostly not sensible, we did not conduct human evaluations on those results. } A licensed patent attorney and an experienced patent engineer, both with extensive experience in drafting patent applications, conduct the evaluation and agree on the results. 

Five criteria are employed to evaluate the quality of the patent claims and details of the rating strategies are listed in Appendix~\ref{humanrating}.
The criteria include: \textbf{1. Completeness of Essential Features:} The extent to which the generated claims encapsulated all critical aspects of the invention. \textbf{2. Conceptual Clarity:} The clarity and unambiguity of the language used in the claims. \textbf{3. Consistency in Terminology:} The uniformity in the use of terms throughout the claims. \textbf{4. Technical Correctness of Feature Linkages:} The accuracy with which the features were interconnected and related. \textbf{5. Overall Quality:} An aggregate measure combining all the above criteria.

\section{Results and Discussion}

\subsection{RQ1: Does description-based claim generation outperform previous abstract-based methods?}

\input{tabs/abstract_description}

Table~\ref{table:comparison_claim_generation} demonstrates the evaluation results of the claims generated based on abstract and description. When feeding descriptions as the input into Llama-3, there is a noticeable improvement across all metrics compared to using abstracts as input. BLEU, R-1, and R-L increase by approximately 12\%, 10\%, and 7\% respectively, showing a higher degree of word and sequence overlap with reference claims. In addition, the BERTScore ascends from 86.17\% to 88.40\%, underscoring a better semantic similarity with the gold claims.

To clearly illustrate the generation differences, we provide example claims outputs in Figure~\ref{fig:abs_claims}. The key features included in the gold claims are highlighted in color. The claim generated from the abstract only comprises some imprecise features. For example, \textit{upper crossbar} mentioned in the original claim is missing and it includes \textit{a frame body} compared to \textit{a generally rectangular frame body} in the referenced claim. In contrast, the claim generated with descriptions as input includes all essential features and uses precise wording. 

We argue that the superiority of description-based generation is because it provides more detailed and expansive information about an invention, while patent abstracts are usually generic and vain. Descriptions delve into patent specifics, including technical details, functionalities, and applications. This richness in detail offers the model more textual cues and a broader context, enabling it to generate more accurate and detailed text.

\textbf{Takeaways } Description-based generation can improve claim quality compared to abstract-based inference. Generated claims include more precise features that are covered in the gold claims, increasing the completeness of outputs. 

\input{tabs/automated_and_human_evaluation}

\input{tabs/main_2}

\subsection{RQ2: Does current patent-specific LLM outperform general LLMs?}

Table~\ref{table:merged_evaluation} presents a comparison of the performance metrics for ten different models. Although PatentGPT-J is specialized for patent texts, it shows significantly lower scores across all metrics. Additionally, our manual examination of the claims generated by PatentGPT-J revealed that they are often nonsensical and lacking in detail. The claims tend to be short and repetitive, further underscoring the model's limitations.

We argue that the poor performance can be attributed to three possible reasons:  (1) The model may not be pre-trained on enough data and steps so the model does not generalize well on patent texts. PatentGPT-J is trained on 11 billion tokens, while Llama-3 is trained on over 15 trillion tokens. (2) The model has not been instruction-tuned \citep{ouyang2022training}. Instruction tuning involves fine-tuning a pre-trained model on a dataset that includes specific instructions or guidance along with the input data. Without this tuning phase, the model may struggle to comprehend and respond to our target purposes. (3) Another issue is the limited context length of PatentGPT-J. The maximum length of text for context (prompt) plus generated text length is 2,048 tokens, which is far below the average document length of 5,451 plus claim length of 962. 

Additionally, SaulLM performs slightly better than PatentGPT-J but still substantially worse than others. Based on our observation, the outputs of SaulLM can be categorized into the following cases: (1) The generated texts have a similar format to patent claims. This indicates that the legal-specific LLM can somehow capture the structure and format of patent claims. Although such outcomes are rare, they demonstrate the potential of legal-specific LLMs on patent claim generation. A possible reason is that SaulLM has been trained on structured patent texts, allowing it to replicate this format in its outputs. (2) The output is a short narrative paragraph of what the patent has claimed. This result shows the model's tendency towards narrative text instead of structured patent claims. This tendency can be attributed to the model’s training predominantly on narrative and explanatory legal content, rather than strictly formatted patent claims. (3) The model generates a summary of the patent description. It suggests that the model might have misunderstood the input prompts. The instruction-tuning process of SaulLM may not be sufficient or the model tends to recognize patent claims as short summaries.  (4) The output makes no sense or repeats the same sentence multiple times. This model's inability may be due to insufficient training and data. Whereas Llama-3 is pre-trained on 15 trillion tokens, SaulLM is based on about 30 billion tokens.

\textbf{Takeaways } (1) PatentGPT-J demonstrates poor performance on claim generation, indicating that current patent-specific LLMs are far from satisfactory. We suggest research groups with enough resources can pre-train such models for this important field. (2) SaulLM designed for the legal domain also under-performs compared to other general LLMs.  Although the patent is one type of legal document, it features specific language and format requirements that differentiate it from others. Hence, LLMs for the general legal domain may not be well-suited for patent claim generation.

\subsection{RQ3: How do LLMs perform on patent claims’ completeness, clarity, consistency, and logical linkage?}

\noindent \textbf{Feature Completeness} In most cases, each first claim generated by LLMs is found to be error-free and accurately reflects the reference claims and the description. This finding aligns with the automated evaluation results listed in Table~\ref{table:evaluation_claims}. The scores of the first claim on all metrics are largely higher than the results on the remaining claims. This accuracy is likely because the description often includes the independent claim verbatim before introducing additional optional features and embodiments. Consequently, the model's ability to extract and replicate this claim correctly is understandable, given the straightforward nature of the task. 

However, the model's ability to generate the subsequent dependent claims is much worse. Llama-3 substantially fails to capture the features claimed in the referenced dependent claims. Three main weaknesses are raised in the usage of Llama-3. (1) Llama-3 produces claims that merely reference another claim without specifying additional features, rendering these claims redundant and devoid of any added value. (2) Llama-3 sometimes misclassifies the feature types. For example, Llama-3 tends to misclassify device-specific features as procedural features, using active formulations. This misclassification is a serious flaw as it contradicts standard practice and would likely lead to objections during patent examination. (3) Llama-3 generates some repetitive claims, where exactly the same claims are continuously generated multiple times. 

We find two key improvements of completeness on problems (1) and (2) from Llama-3-FT. Some dependent claims include additional features that provide some level of limitation or specificity to the respective independent claim it referenced. Moreover, Llama-3-FT rarely misclassifies features, representing a significant improvement in adhering to standard practice. Therefore, fine-tuning improves the completeness scores from 4.0 to 5.3, but significant gaps still remain. The feature coverage is not good enough and the fine-tuned models also generate repetitive claims. Human evaluation aligns with automated evaluation, where the fine-tuning of Llama-3 improves on all automated metrics compared to its zero-shot performance, especially about an 8.5\% gain in R-L. This suggests fine-tuning leads to more lexical overlaps and semantic similarity with referenced claims. Meanwhile, larger models can also bring this type of improvement. Llama-3-70B increases the R-L from 38.88\% to 41.79\% compared to the 8B version, and Mistral-8$\times$22B improves from 43.67\% to 46.30\% versus its 8$\times$7B version. 

Additionally, the features claimed in referenced dependent claims are more comprehensively covered by larger models, such as Mixtral-8$\times$7B with a completeness score of 5.7, although not all features are well represented. Furthermore, GPT-4 avoids outputting repetitions of claim features, which is an improvement compared to Llama-3-based models.

\noindent \textbf{Conceptual Clarity }
Llama-3-FT also improves conceptual clarity from 4.8 to 5.8, which indicates the claims are mostly clear but contain some ambiguous language. 
The main problem is the incorrect use of pronouns, such as "it". The use of such pronouns must not introduce ambiguity. Clarity and precision are paramount in patent claims to ensure that the scope of the invention is well-defined and legally enforceable. Therefore, it is essential to make sure any references, including pronouns, are unmistakably clear when drafting patent claims. Preferably, pronouns are avoided entirely. 

Although Llama-3-MFT remains advantages brought by fine-tuning, MFT introduces new problems in conceptual clarity, decreasing the score from 5.8 to 4.9. Llama3-MFT sometimes produces claims not entirely supported by the description, incorporating features with no practical sense for the protection scope or potential differentiation. Furthermore, the claim set generated by Llama3-MFT suffers from several structural and technical issues: The use of unclear terms, such as "desired", contributes to the ambiguity and potential legal challenges. These issues significantly reduce the human rating of language clarity.  Similarly, Llama-3-MFT does not improve automated evaluation results either. Compared to single-task fine-tuning, Llama-3-MFT achieves almost the same BLEU and BERTScore, while there are approximately 1\% performance drops on R-1 and R-L. 

This result implies that despite the similarity between claim generation and patent summarization, they may not be suitable for multi-task learning. Although they both require understanding complex and long documents to generate shorter target texts, the training objectives are in conflict. Patent abstracts are general, while claims include technical details and are on average about seven times longer than the abstract. Moreover, the requirements of precision and language unambiguity for the abstract are not as high as in claims. These conflicts can make the model unable to optimally learn the specific pattern for each task, thus compromising performance.

\noindent \textbf{Terminology Consistency } All models achieve 6.5 and higher scores on terminology consistency, indicating that the terminology is largely consistent with minor inconsistencies. This means that current LLMs can mostly correctly recognize and use domain-specific terminology. However, even minor inconsistency would result in the rejection and/or loss of value of a patent application. Therefore, more work is needed to ensure terminology is completely consistent throughout the claims. 

\noindent \textbf{Correctness of Feature Linkage }
Llama-3 shows a low score of 4.1 on the correctness of feature linkage. The features in dependent claims appear to be randomly selected from the description, without considering the importance or technical relevance of the features. In addition, Llama-3 tends to formulate separate claims instead of summarizing them into one dependent claim. The result suggests the model lacks the technical understanding necessary to evaluate the importance of technical features, their interaction, and their significance in the overall context of invention. Fine-tuning partially fixes these problems, improving the score to 5.3. 

In contrast, GPT-4 uniquely groups alternative embodiments within dependent claims logically, rather than random selection and describing them as separate claims. It enhances the technical coherence and readability of the claims, improving the human evaluation score to 5.9. Furthermore, GPT-4 provides a cohesive and structured claim set, instead of generating multiple repetitive claims. 

\noindent \textbf{Overall Quality }
GPT-4 shows relatively low scores on BLEU, R-1, and R-L, suggesting that the output covers fewer word sequences in the referenced claims compared to other models. These low scores are probably a consequence of the trend that claims generated by GPT-4 are much shorter and more concise. We report the detailed statistics of generated claims in Appendix~\ref{statistic}. The result demonstrates that the outputs of GPT-4 only have 482 tokens and 6.56 claims on average, which are significantly less than others. 

Nonetheless, GPT-4 results in the best human evaluation. The primary claim is mostly accurate, and the dependent claims exhibit better technical relevance, clarity, and logical grouping. The model demonstrates a better understanding of technical features, their interactions, and their significance within the overall context of the invention, indicating its capability to evaluate the entire text and generate a coherent claim set. Due to the large inconsistency between human and existing automatic evaluation methods, a promising research direction is to develop new automated metrics that have closer alignment to human evaluation, such as using LLM-based evaluators \citep{liu-etal-2023-g, jiang2025towards}. 

Despite the improvements, the claim set generated by GPT-4 still requires some revisions to ensure the claims' legal and technical robustness. Adjustments are necessary to identify and eliminate features that do not contribute to differentiation, known as \textit{null features}. Some restrictions in dependent claims are not optimal compared to other possible restrictions, requiring further refinement.

\noindent \textbf{Takeaways } 
(1) All models can generate a high-quality first independent claim, but perform poorly on dependent claims. (2) Fine-tuning brings two major improvements in feature completeness, avoiding replicating claims without adding additional features and avoiding misclassifying features. 
(3) Although MFT keeps the improvement brought by fine-tuning, new problems are introduced, significantly decreasing conceptual clarity. 
(4) Claims generated by all models demonstrate relatively high terminology consistency, but further improvements are needed. 
(5) Larger models can select and group features within dependent claims logically, instead of randomly choosing features and describing them separately. 
(6) Although GPT-4 generates short texts and has less sequence coverage with referenced claims, it achieves a higher rating in all human evaluation metrics. 
(7) Despite promising, there are challenges unsolved, particularly in achieving the completeness of essential features and maintaining technical accuracy.

\section{Conclusion}
We construct a dataset for description-based patent claim generation and evaluate various LLMs. Our results indicate that description-based claim generation surpasses previous abstract-based methods. We also observe that current patent-specific or legal-specific LLMs exhibit inadequate performance, underscoring the need for further research.
Moreover, while LLMs can output high-quality first independent claims, their effectiveness significantly diminishes with subsequent dependent claims. 
Additionally, fine-tuning can enhance feature completeness, conceptual clarity, and feature linkage, whereas multi-task fine-tuning reduces conceptual clarity. 
GPT-4 outperforms other models across all human evaluation metrics, demonstrating better feature coverage, conceptual clarity, and technical coherence. Despite promising, the models are not yet satisfactory for practical applications.

\section*{Limitations}
We acknowledge several limitations of our research. 
Firstly, we contribute pioneering work on patent claim generation, focusing on patent descriptions limited to fewer than 8,000 tokens. Our results show that LLMs generally underperform with documents of such length. Since longer documents may pose more challenges for LLMs, our conclusions are not compromised. 
Secondly, our investigation is restricted to English-language patents. Future work may focus on patents in other languages.
Thirdly, the effects of hyperparameter tuning are not explored in this research. More appropriate hyperparameters for training and inference may improve the results. 

To enhance the quality of generated claims, future research could explore recent advancements in uncertainty estimation for LLMs \citep{zhang-etal-2024-luq, zhang2024atomic, yang2024logu}. By identifying uncertain generations, human verification can refine the output and improve performance beyond fully automated generation.

\section*{Ethics Statement}
Llama-3 is under \textit{META LLAMA 3 COMMUNITY LICENSE AGREEMENT}. No license is found for PatentGPT-J. GPT-4 is under a commercial license provided by OpenAI, and we access it through its API. The HUPD dataset uses the \textit{CC-BY-SA-4.0} license, and our derived HUPD-DCG dataset will use the same license as HUPD. Since HUPD is publicly available, we do not further check potential personal information and offensive content. The use of existing artifacts is consistent with their intended use. Our proposed dataset is used for description-based claim generation and is compatible with the original access conditions.


\bibliography{anthology,custom}

\appendix

\section{Patent Description, Abstract, and Claims}
\label{examplepatent}

\input{tabs/example_patent}

Table~\ref{table:example_patent} shows an example of patent description, abstract, and claims. The patent \textit{description} provides a comprehensive illustration of the invention, including structural and functional properties of the invention based on detailed embodiments and examples as well as the context in which it is used. The \textit{abstract} is a short summary that provides a quick overview of the technical invention in the patent document. It is usually limited to a few sentences and must be concise, covering the main technical aspects without going into details. Patent \textit{claims} are the legally significant part of a patent application, defining the technical scope of the invention to withstand legal scrutiny and secure legal protection. Claims must be crafted with high-level precision and clarity, comprising technological aspects that warrant protection and include the features that constitute the invention. 

Patent claims can be further classified into two categories: \textit{independent claims} and \textit{dependent claims}. Independent claims outline the essential features of an invention, without reference to other claims. They aim to encompass the invention as broadly as feasible, covering ideally all reasonable implementations and modifications, yet are specific enough to distinguish the invention from existing technologies. Dependent claims, which are attached to an independent claim, incorporate additional features or limitations. These serve to refine and specify particular embodiments or variations of the invention, enhancing the details and scope of patent protection.

\section{Model Details}
\label{modeldetails}
\noindent \textbf{PatentGPT-J } We select the 6B version of PatentGPT-J \citep{lee2023evaluating}, which is pre-trained from scratch based on GPT-J-6B model architecture \citep{gpt-j}. This is currently the largest open-source LLM specifically designed for patents. The dataset used for pre-training includes patent titles, abstracts, claims, and descriptions of patent documents ranging from 1976 to 2021. Overall, PatentGPT-J is trained on 11 billion tokens for 350 thousand steps. 

\noindent \textbf{GPT-4 } We use the latest GPT-4 model \citep{achiam2023gpt}, representing the state-of-the-art LLM with substantial general knowledge and outstanding reasoning capabilities.\footnote{gpt-4-turbo-2024-04-09: \url{https://platform.openai.com/docs/models/gpt-4-and-gpt-4-turbo}} Notably, GPT-4 shows remarkable performance on summarization tasks \citep{pu2023summarization}. Hence, we do not evaluate other LLMs specifically trained for summarization. 

\noindent \textbf{Llama-3 }  Among open-source LLMs, we choose the recent Llama-3, pre-trained on over 15 trillion tokens of data from publicly available sources.\footnote{\url{https://llama.meta.com/llama3/}} Llama-3 outperforms most open-source models on common industry benchmarks. We opt for both the Llama-3-8B-Instruct and Llama-3-70B-Instruct versions because they have been fine-tuned to execute user instructions more accurately and effectively. 

\noindent \textbf{Llama-3-FT } Llama-3-FT is a fine-tuned version of the original Llama-3-8B-Instruct model based on our train set. We use LoRA \citep{hu2021lora} for training, a parameter-efficient approach to reduce computational needs while maintaining comparable performance. The inputs are instruction prompts and patent descriptions, while the output is corresponding patent claims. Appendix~\ref{experimentsdetail} lists experimental details. 

\noindent \textbf{Llama-3-MFT } Multi-task fine-tuning (MFT) has shown effectiveness in many applications, such as code generation \citep{liu2023mftcoder}. MFT trains the model on multiple tasks simultaneously to enhance its ability to better generalize across a broader range of tasks by learning shared and task-specific features. Our Llama-3-MFT model is trained on two tasks, description-based claims generation and summarization (abstract generation). Both tasks require processing complex documents and generating shorter texts. Thus, we aim to investigate whether MFT could bring extra benefits to claim generation compared to single fine-tuning. 

\noindent \textbf{Mistral } Mistral is an open-sourced LLM designed for superior performance and efficiency, which outperforms Llama-2-13B on some tasks \citep{jiang2023mistral}. We use the version of Mistral-7B-Instruct-v0.3 that has been instruction-tuned to follow users' prompts.\footnote{\url{https://huggingface.co/mistralai/Mistral-7B-Instruct-v0.3}}

\noindent \textbf{SaulLM } SaulLM-7B is designed for the legal domain, which adopts the Mistral-7B architecture and is trained on an English legal corpus of over 30 billion tokens \citep{colombo2024saullm}. The dataset includes 4.7B tokens of patent documents from the United States Patent and Trademark Office (USPTO). We use the SaulLM-7B-Instruct version for experiments. 

\noindent \textbf{Mixtral }  Mixtral-8$\times$7B is based on Sparse Mixture of Experts (SMoE) \citep{jiang2024mixtral}.  It has the same architecture as Mistral-7B, but each layer comprises 8 feedforward blocks (i.e. experts). A router network in each layer selects two experts to process the current state and combine the outputs. This model only uses 13B active parameters during inference, but outperforms the 70B version of Llama-2. We evaluate both Mixtral-8$\times$7B-Instruct and Mixtral-8$\times$22B-Instruct. 

\input{tabs/claim_statistics}

\input{tabs/prompt_results}

\section{Experimental Details}
\label{experimentsdetail}
All fine-tuning and inference processes are conducted on NVIDIA A100 GPUs. The total running time is about 700 hours. We use the first 90\% documents of the entire train set as the training dataset and the remaining 10\% as the validation set.  During fine-tuning, we use a LoRA rank of 96, LoRA alpha of 32, dropout of 0.05, batch size of 1, learning rate of 5e-5, weight decay of 0.1, and training epochs of 5. The training will stop early if the loss does not decrease after 500 steps. For inference, we set the temperature to 0.1 and the maximum generation tokens to 1,024. We have employed a standard prompt format to maintain consistency and focus on assessing the capabilities and limitations of different LLMs in this domain. The prompt is: \textit{You are a patent expert. Given the following patent description, write patent claims.} For automated evaluation metrics, we use the package from the HuggingFace Evaluate library.\footnote{\url{https://github.com/huggingface/evaluate}}

\section{More Results}

\subsection{Statistics of Generated Claims}
\label{statistic}

To explore the attributes of outputs generated by various models, we conducted a detailed statistical analysis. This analysis encompasses the average number of tokens, number of claims, length of claims, structural complexity, and readability. Structural complexity is assessed based on the ratio of subordinate clauses to the total number of sentences. For readability, we employ the Flesch-Kincaid Grade Level formula \citep{kincaid1975derivation}, where a lower score represents easier readability.

Patent texts have unique requirements to pass rigorous patent scrutiny. For example, the precision requirement and information density of patent texts are higher than in everyday language. The patent language focuses more on precision and accuracy instead of readability. Such precision requirement typically leads to high repetitiveness in both terminology and structure of sentences, paragraphs, and sections. Furthermore, sentences are often overburdened because they use relative or adverbial clauses to include specifications for precision or add examples for a wider scope. LLMs, such as GPT-4, which is trained predominantly on more accessible texts, naturally tend towards generating more readable content. Thus, this difference lets LLMs output more readable text than the gold claims, but it is irrelevant to the human expert assessment, which aims at formal and legal aspects.

\subsection{Evaluation on Alternative Prompts}

To investigate the influence of different prompts, we have applied an alternative prompt to this task while the other settings remain unchanged: \textit{You are a patent expert. Given the following patent description, generate patent claims. Ensure the claims include all essential features, the language is unambiguous, the terminologies are used consistently, and features are interconnected and related accurately.} Table \ref{table:evaluation_prompt} shows that all models perform better or maintain similar scores when using this revised prompt. Interestingly, the alternative prompt only brings marginal improvements to SaulLM-7B, Llama-3-8B, Llama-3-70B, and Mixtral-8$\times$7B, whereas Mistral-7B and GPT-4 are more sensitive to different prompts. 

\input{tabs/ipc_results}

\subsection{Evaluation on IPC Labels}

We report the automated evaluation results of each patent category at the International Patent Classification (IPC) section level in Table \ref{table:evaluation_ipc}. We select four of the most frequent sections, including H (35\%): Electricity; G (25\%): Physics; B (14\%): Performing operations,
Transporting; A (12\%): Human necessities; and others (14\%). Table \ref{table:evaluation_ipc} reveals that all models exhibit fluctuations in performance across different IPC sections. For example, the performance of each metric of the same model in section B is consistently higher than others. In contrast, models usually perform worse on patents of section A. The fluctuations can be more than 15\%, for instance, Mistral-8$\times$7B achieves 48.16\% R-L in section B, but only 32.39\% in section A. The section-wise variance suggests that certain sections may be inherently more challenging for models to handle, possibly due to the technical complexity or uniqueness of the language used in those sections. Overall, the results highlight the challenges in achieving uniformly high performance across different types of patents, suggesting the need for improvements to enhance model robustness and adaptability to diverse content types.

\subsection{Evaluation on Abstracts}

The purpose of this research is to evaluate model performance on claim generation, but we also report the results of summarization in Table~\ref{table:evaluation_abstract}. The scores on abstract are better than those on claims, indicating that claim generation is more challenging. We find a similar trend of abstract generation: fine-tuning can improve performance, whereas multi-talk fine-tuning cannot bring extra benefits. 

\input{tabs/abstract_evaluation}

\section{Human Rating Strategies}
\label{humanrating}

Patent professionals are given the patent description, referenced claims, and claims generated by Llama-3, Llama-3-FT, Llama-3-MFT, and GPT-4. They evaluate each automated generated claim based on the following details of human evaluation criteria, illustrated in Table~\ref{tab:humanrating}. They were informed that the average rating would be used as human evaluation results in the paper, and no ethics review board was involved.

\input{tabs/annotation_criteria}

\end{document}

%% file: tabs/comparison_with_summarization.tex
\begin{table*}[!ht]
\centering
\resizebox{\textwidth}{!}{
\begin{tabular}{lp{8cm}|p{8cm}}
\toprule
\textbf{} & \multicolumn{1}{c}{\textbf{Claim Generation (claims)}} & \multicolumn{1}{|c}{\textbf{Summarization (abstract)}} \\
\midrule
\textbf{Similarity} & \multicolumn{2}{p{15cm}}{Require understanding of long and complex documents. Output shorter texts with target information.} \\
\midrule
\textbf{Goal} & Define the legal protection scope of inventions & Help readers quickly grasp the invention's innovation \\
\midrule
\textbf{Content} & Detailed features of the invention & A concise summary of the invention \\
\midrule
\textbf{Language} & Strict writing standard with precise and clear wording & More accessible and narrative language \\
\midrule
\textbf{Structure} & An ordered list of independent and dependent claims & A coherent paragraph \\
\midrule
\textbf{Audience} & Patent professionals (e.g., examiners, lawyers) & Broader audience (e.g., researchers, investors) \\
\bottomrule
\end{tabular}
}
\caption{Comparison of claim generation and summarization. Patent claim generation is more difficult because it requires more details, higher language precision, and logical structure. }
\label{table:comparison}
\end{table*}

%% file: tabs/patent_datasets.tex
\begin{table*}[!ht]
\centering
\footnotesize
\begin{tabular}{lccc}
\toprule
\textbf{Dataset} & \textbf{Task} & \textbf{Details} & \textbf{Size} \\
\midrule
\citet{casola2023creating} & Simplification & Original sentence $\rightarrow$ simple sentence & 200K \\
GED \citep{wirth-etal-2023-building} & Translation & FR, DE $\leftrightarrow$ EN & 19K \\
EuroPat \citep{heafield-etal-2022-europat} & Translation & DE, ES, FR, HR, NO, PL $\leftrightarrow$ EN & 51M \\
BigPatent \citep{sharma-etal-2019-bigpatent} & Summarization & Description $\rightarrow$ abstract & 1.3M \\
Patent-CR \citep{jiang-etal-2025-patent} & Claim revision & Claim $\rightarrow$ claim & 22K \\ 
\rowcolor{gray!30}
HUPD-DCG (Ours) & Patent claim generation & Description $\rightarrow$ claim & 9.5K \\
\bottomrule
\end{tabular}
\caption{Publicly available datasets for patent text generation. Abbreviations: FR - French, DE - German, ES - Spanish, HR - Croatian, NO - Norwegian, PL - Polish, EN - English.}
\label{table:datasets}
\end{table*}

%% file: tabs/dataset_statistics.tex
\begin{table*}[ht!]
\footnotesize
\centering
\begin{tabular}{lcccccc}
\toprule
\textbf{} & \multicolumn{3}{c}{\textbf{Train}} & \multicolumn{3}{c}{\textbf{Test}} \\
\cmidrule(lr){2-4} \cmidrule(lr){5-7}
 & \textbf{Description} & \textbf{Abstract} & \textbf{Claims} & \textbf{Description} & \textbf{Abstract} & \textbf{Claims} \\
\midrule
Number of documents & 8244 & 8244 & 8244 & 1311 & 1311 & 1311 \\
Average document length & 5491 & 133 & 969 & 5451 & 134 & 962 \\
Average number of sentences & 179 & 3.5 & 12.4 & 179 & 3.6 & 11.4 \\
Total words & 704 & 55 & 129 & 715 & 55 & 128 \\
\midrule
Coverage & - & 0.08 & 0.17 & - & 0.08 & 0.16 \\
Compression ratio & - & 48 & 7.6 & - & 48 & 8.3 \\
\bottomrule
\end{tabular}
\caption{Data statistics of our HUPD-DCG dataset. Train and test sets have similar attributes, showing the dataset's validity. Claims are more detailed and more than seven times longer than abstracts. }
\label{table:dataset_statistics}
\end{table*}

%% file: tabs/abstract_description.tex
\begin{table}[t!]
\footnotesize
\centering
\renewcommand{\arraystretch}{0.1} 

\begin{tabular}{
    p{1.6cm} 
    >{\centering\arraybackslash}p{1cm} 
    >{\centering\arraybackslash}p{0.75cm} 
    >{\centering\arraybackslash}p{0.75cm} 
    >{\centering\arraybackslash}p{1.5cm}
}
\toprule
\textbf{Model} & \textbf{BLEU} & \textbf{R-1} & \textbf{R-L} & \textbf{BERTScore} \\
\midrule
\makecell[l]{GPT-2-FT\\(abstract) 
 \\ \cite{lee2020controlling}
} & - & 52.38 & - & - \\

\midrule

\makecell[l]{Llama-3 \\ (abstract)} & 21.98 & 46.30 & 31.38 & 86.17 \\

\midrule

\makecell[l]{Llama-3\\(description)} & \textbf{34.32} & \textbf{56.72} & \textbf{38.44} & \textbf{88.40} \\

\bottomrule
\end{tabular}
\caption{Comparison of abstract-based and description-based claim generation. 
Note that the dataset used in the previous study is not publicly available, so we list previous results as a reference, which are not directly comparable to our experiments and results. This table showed that the description-based claim generation outperformed previous abstract-based methods.}
\label{table:comparison_claim_generation}
\end{table}

%% file: tabs/automated_and_human_evaluation.tex
\begin{table*}[ht!]
\footnotesize
\centering
\resizebox{0.99\linewidth}{!}{
\begin{tabular}{lccccccccccc}
\toprule
& \multicolumn{5}{c}{\textbf{Human Evaluation}} & \multicolumn{4}{c}{\textbf{Automated Evaluation}}  \\
\cmidrule(lr){2-6} \cmidrule(lr){7-10}
\textbf{Model} & \makecell{\textbf{Feature} \\ \textbf{Completeness}} & \makecell{\textbf{Conceptual} \\ \textbf{Clarity}} & \makecell{\textbf{Terminology} \\ \textbf{Consistency}} & \makecell{\textbf{Correctness of} \\ \textbf{Feature Linkage}} & \makecell{\textbf{Overall} \\ \textbf{Quality}} & \textbf{BLEU} & \textbf{R-1} & \textbf{R-L} & \textbf{BERTScore} \\
\midrule
\multicolumn{3}{l}{\textbf{Domain-specific LLMs}} \\
PatentGPT-J-6B  & - & - & - & - & - & 12.86 & 30.68 & 23.07 & 80.24 \\
SaulLM-7B  & - & - & - & - & - & 12.68 & 36.63 & 25.10 & 83.13\\
\midrule

\multicolumn{3}{l}{\textbf{Base LLMs}} \\
Mistral-7B  & 5.0 & 4.3 & 4.6 & 5.0 & 4.8 & 29.70 & 49.17 & 36.20 & 85.33\\
Llama-3-8B & 4.0 & 4.8 & 6.6 & 4.1 & 4.6 & 35.42 & 57.17 & 38.88 & 88.49 \\
\midrule

\multicolumn{3}{l}{\textbf{Fine-tuned LLMs}} \\
Llama-3-FT  & 5.3 & 5.8 & 6.9 & 5.3 & 5.6 & \textbf{37.52} & 59.96 & \textbf{47.35} & \textbf{89.45}\\
Llama-3-MFT & 5.1 & 4.9 & 6.5 & 5.3 & 5.4 & 37.27 & 58.86 & 46.25 & 89.41 \\
\midrule

\multicolumn{3}{l}{\textbf{Large-sized LLMs}} \\
Llama-3-70B & 5.6 & 5.0 & 5.2 & 5.0 & 5.3 & 36.40 & 59.89 & 41.79 & 87.44 \\
Mixtral-8$\times$7B & \textbf{5.7} & 5.0 & 5.0 & 5.0 & 5.2 & 37.03 & 60.18 & 43.67 & 88.51 \\
Mixtral-8$\times$22B & 5.6 & 4.8 & 5.4 & 5.2 & 5.3 & 33.96 & \textbf{60.57} & 46.30 & 88.97 \\
GPT-4 &  5.4 & \textbf{6.3} & \textbf{7.4} & \textbf{5.9} & \textbf{6.0} & 15.73 & 52.59 & 37.73 & 87.34  \\
\bottomrule
\end{tabular}}
\caption{Automated and human evaluation results. The best scores for each metric are marked in \textbf{bold}. PatentGPT-J and SaulLM show poor performance and sometimes nonsensical outputs. Fine-tuning improves on both automated and human evaluation metrics. Multi-task fine-tuning performs slightly worse than single fine-tuning.  GPT-4 demonstrates lower automated evaluation scores but has the best overall quality based on human evaluation. }
\label{table:merged_evaluation}
\end{table*}

%% file: tabs/main_2.tex
\begin{table*}[ht!]
\footnotesize
\centering
\resizebox{0.85\linewidth}{!}{
\begin{tabular}{lcccccccc}
\toprule
 & \multicolumn{4}{c}{\textbf{First Claim}} & \multicolumn{4}{c}{\textbf{Remaining Claims}} \\
\cmidrule(lr){2-5} \cmidrule(lr){6-9}
\textbf{Model} & \textbf{BLEU} & \textbf{R-1} & \textbf{R-L} & \textbf{BERTScore} & \textbf{BLEU} & \textbf{R-1} & \textbf{R-L} & \textbf{BERTScore} \\
\midrule
Mistral-7B & 50.53 & 64.69 & 57.44 & 90.19 & 29.24 & 49.44 & 36.03 & 85.52 \\
Llama-3-8B & 36.93 & 60.50 & 51.61 & 90.58 & 29.24 & 51.46 & 33.68 & 87.11 \\
Llama-3-FT & 49.78 & 66.04 & 59.24 & 91.86 & 32.45 & 54.59 & 42.58 & 87.86 \\
Llama-3-MFT & 48.95 & 65.70 & 58.72 & 91.77 & 31.56 & 52.96 & 41.11 & 87.48 \\
Llama-3-70B & 43.12 & 65.39 & 57.76 & 90.35 & 33.18 & 54.62 & 36.99 & 86.50 \\
Mixtral-8$\times$7B & 46.18 & 65.04 & 57.58 & 90.44 & 31.15 & 55.75 & 38.54 & 87.37 \\
Mixtral-8$\times$22B & 48.36 & 66.46 & 59.61 & 91.59 & 26.59 & 52.78 & 39.94 & 87.49 \\
GPT-4 & 33.73 & 59.29 & 48.49 & 88.51 & 7.47 & 42.87 & 30.41 & 86.51 \\
\bottomrule
\end{tabular}}
\caption{Automated evaluation results on the first independent claim and remaining claims. The scores of all metrics for the first claim are significantly higher than the counterparts of the remaining claims. }
\label{table:evaluation_claims}
\end{table*}

%% file: tabs/example_patent.tex
\begin{table*}[t]
\centering
\footnotesize
\begin{tabular}{p{15.5cm}}
\toprule
\textbf{Patent description:} \\
Embodiments of the present invention will be described in detail below. In the present specification, the same or equivalent components are designated by the same reference numerals. First Embodiment FIG. 1A is a cross-sectional view illustrating a structure of a semiconductor light-emitting element (hereinafter, may be referred to simply as a light-emitting element or an element) 10 according to a first embodiment. The semiconductor light-emitting element 10 has a structure in which a semiconductor structure layer SS is formed on a mounting substrate (hereinafter, may be referred to simply as a substrate) 11 formed from sapphire. The semiconductor structure layer SS will be concretely described below. An n-type semiconductor layer 12 serving as a first semiconductor layer is formed from, for example, a GaN layer containing an n-type dopant (for example, Si).   \\
\textit{(The remaining description is omitted)} \\
\midrule
\textbf{Patent abstract: } \\
A semiconductor light-emitting element according to the present invention includes a first semiconductor layer of a first conductivity type, a light-emitting functional layer formed on the first semiconductor layer, and a second semiconductor layer that is formed on the light-emitting functional layer and is of a second conductivity type opposite to that of the first semiconductor layer. The light-emitting functional layer includes a doped layer that is formed on the first semiconductor layer and doped with a dopant of the second conductivity type, a base layer formed on the doped layer, the base layer having such a composition that causes stress and strain in said base layer from the doped layer, said base layer including a plurality of base segments formed in a random net shape, and a quantum well structure layer formed on the base layer. \\
\midrule
\textbf{Patent claims: } \\
1. A semiconductor light-emitting element comprising: a first semiconductor layer of a first conductivity type; a light-emitting functional layer formed on said first semiconductor layer; and a second semiconductor layer that is formed on said light-emitting functional layer and is of a second conductivity type opposite to that of said first semiconductor layer, wherein said light-emitting functional layer includes: a doped layer that is formed on said first semiconductor layer and doped with a dopant of said second conductivity type, a base layer formed on said doped layer, said base layer having such a composition that causes stress and strain in said base layer from said doped layer, said base layer including a plurality of base segments formed in a random net shape, and a quantum well structure layer formed on said base layer. \\
2. The semiconductor light-emitting element according to claim 1, wherein said quantum well structure layer is an undoped layer. \\
3. The semiconductor light-emitting element according to claim 1, wherein said dopant of said doped layer is Mg. \\
4. The semiconductor light-emitting element according to claim 1, wherein: said first semiconductor layer and said doped layer include a GaN composition; said quantum well structure layer includes a quantum well layer and a barrier layer that are formed on said base layer; said base layer and said barrier layer include an AlN or AlGaN composition; and said quantum well layer includes an InGaN composition. \\
5. The semiconductor light-emitting element according to claim 1, further comprising a second light-emitting functional layer between said doped layer and said first semiconductor layer, said second light-emitting functional layer including a uniformly flat quantum well structure. \\
6. The semiconductor light-emitting element according to claim 5, wherein said second light-emitting functional layer has a center emission wavelength different from that of said quantum well layer. \\    
\bottomrule
\end{tabular}
\caption{Example patent description, abstract, and claims of patent US20180062039A1. The patent description describes in detail the technical aspects of the invention. The abstract is a brief summary of the invention. Claims define the legal boundaries of the patent.}
\label{table:example_patent}
\end{table*}

%% file: tabs/claim_statistics.tex
\begin{table*}[ht!]
\footnotesize
\centering
\begin{tabular}{lccccc}
\toprule
\textbf{Model} & \makecell{\textbf{Num. of} \\ \textbf{tokens}} & \makecell{\textbf{Num. of} \\ \textbf{claims}} & \makecell{\textbf{Avg. claim} \\ \textbf{length}} & \makecell{\textbf{Structure} \\ \textbf{complexity}} & \makecell{\textbf{Readability} \\ ($\downarrow$)} \\
\midrule
Gold standard & 962 & 11.36 & 84.69 & 1.01 & 31.96 \\
GPT-4 & 482 & 6.59 & 73.05 & 1.05 & 16.52 \\
Llama-3-8B & 990 & 19.20 & 51.55 & 0.62 & 21.57 \\
Llama-3-FT & 836 & 14.49 & 57.70 & 1.09 & 30.40 \\
Llama-3-MFT & 857 & 15.36 & 55.81 & 1.11 & 29.66 \\
Llama-3-70B & 760 & 11.15 & 68.15 & 1.62 & 15.29 \\
Mistral-7B & 868 & 17.06 & 50.89 & 0.90 & 26.84 \\
Mixtral-8$\times$7B & 969 & 13.00 & 74.51 & 0.74 & 25.90 \\
Mixtral-8$\times$22B & 732 & 11.68 & 62.73 & 0.84 & 28.19 \\

\bottomrule
\end{tabular}
\caption{Claim statistics of each model output. Note that lower readability scores demonstrate higher readability.}
\label{table:claim_statistics}
\end{table*}

%% file: tabs/prompt_results.tex
\begin{table*}[ht!]
\footnotesize
\centering
\resizebox{0.85\linewidth}{!}{
\begin{tabular}{lcccccccc}
\toprule
 & \multicolumn{4}{c}{\textbf{Normal Prompt}} & \multicolumn{4}{c}{\textbf{Alternative Prompt}} \\
\cmidrule(lr){2-5} \cmidrule(lr){6-9}
\textbf{Model} & \textbf{BLEU} & \textbf{R-1} & \textbf{R-L} & \textbf{BERTScore} & \textbf{BLEU} & \textbf{R-1} & \textbf{R-L} & \textbf{BERTScore} \\
\midrule
SaulLM-7B & 12.86 & 30.68 & 23.07 & 80.24 & 12.99 & 31.45 & 23.71 & 80.41  \\
Mistral-7B & 29.70 & 49.17 & 36.20 & 85.33 & 37.87 & 59.72 & 43.49 & 87.81 \\
Llama-3-8B & 35.42 & 57.17 & 38.88 & 88.49 & 35.61 & 57.50 & 39.17 & 88.51 \\
Llama-3-70B & 36.40 & 59.89 & 41.79 & 87.44 & 37.71 & 60.43 & 41.89 & 87.86 \\
Mixtral-8$\times$7B & 37.03 & 60.18 & 43.67 & 88.51 & 37.86 & 60.04 & 43.25 & 88.53 \\
GPT-4 & 15.73 & 52.59 & 37.73 & 87.34 & 28.39 & 59.53 & 40.61 & 87.72 \\
\bottomrule
\end{tabular}}
\caption{Automated evaluation results on different prompts. }
\label{table:evaluation_prompt}
\end{table*}

%% file: tabs/ipc_results.tex
\begin{table*}[ht!]
\footnotesize
\centering
\resizebox{0.99\linewidth}{!}{
\begin{tabular}{lcccccccc}
\toprule

\textbf{Metric} & \textbf{Mistral-7B} & \textbf{Llama-3-8B} & \textbf{Llama-3-FT} & \textbf{Llama-3-MFT } & \textbf{Llama-3-70B} & \textbf{Mixtral-8$\times$7B} & \textbf{Mixtral-8$\times$22B} & \textbf{GPT-4} \\
\midrule
\multicolumn{3}{l}{\textbf{BLEU}} \\
H (35\%) & 31.83 & 36.18 & 38.50 & 37.13 & 39.46 & 43.19 & 40.74 & 17.56 \\
G (25\%) & 27.62 & 34.60 & 35.41 & 36.46 & 30.02 & 27.37 & 29.85 & 13.64 \\
B (14\%) & 31.16 & 33.82 & 43.03 & 41.71 & 41.21 & 43.76 & 38.49 & 19.73 \\
A (12\%) & 24.74 & 31.76 & 31.39 & 31.96 & 25.87 & 21.04 & 18.08 & 10.91 \\
Others (14\%) & 30.77 & 32.88 & 39.07 & 39.73 & 38.59 & 36.86 & 33.73 & 15.85 \\
\midrule
\multicolumn{3}{l}{\textbf{R-1}} \\
H (35\%) & 50.69 & 58.20 & 62.09 & 59.86 & 60.31 & 64.10 & 62.79 & 54.42 \\
G (25\%) & 47.13 & 56.88 & 58.33 & 57.79 & 59.03 & 57.09 & 58.77 & 51.03 \\
B (14\%) & 49.66 & 57.09 & 63.80 & 62.11 & 62.48 & 64.90 & 62.50 & 56.52 \\
A (12\%) & 46.44 & 54.47 & 53.88 & 53.91 & 53.26 & 48.80 & 51.27 & 46.34 \\
Others (14\%) & 50.83 & 56.36 & 59.19 & 59.07 & 61.75 & 61.05 & 58.97 & 52.07 \\
\midrule
\multicolumn{3}{l}{\textbf{R-L}} \\
H (35\%) & 37.54 & 39.37 & 49.44 & 47.74 & 42.84 & 46.91 & 48.89 & 40.23 \\
G (25\%) & 35.16 & 38.73 & 46.46 & 45.65 & 40.17 & 41.01 & 45.51 & 36.40 \\
B (14\%) & 37.45 & 38.72 & 50.65 & 49.29 & 43.88 & 48.16 & 48.29 & 40.28 \\
A (12\%) & 32.08 & 35.65 & 39.90 & 39.00 & 33.75 & 32.39 & 33.95 & 31.23 \\
Others (14\%) & 36.81 & 37.97 & 46.87 & 46.52 & 44.16 & 42.75 & 45.98 & 36.71 \\
\midrule
\multicolumn{3}{l}{\textbf{BERTScore}} \\
H (35\%) & 85.73 & 88.74 & 90.00 & 89.83 & 87.83 & 89.16 & 89.59 & 87.78 \\
G (25\%) & 84.79 & 88.27 & 89.16 & 89.17 & 86.87 & 88.17 & 88.85 & 86.88 \\
B (14\%) & 85.81 & 88.97 & 90.43 & 90.18 & 88.09 & 89.42 & 89.19 & 87.94 \\
A (12\%) & 84.35 & 87.50 & 87.42 & 87.85 & 85.59 & 85.54 & 86.49 & 86.29 \\
Others (14\%) & 85.59 & 88.56 & 89.34 & 89.32 & 87.95 & 88.49 & 88.97 & 87.22 \\

\bottomrule
\end{tabular}}
\caption{Automated evaluation results of each patent category at the International Patent Classification (IPC) section level. Four of the most frequent sections are selected (H: Electricity. G: Physics. B: Performing operations; Transporting. A: Human necessities).   The scores of the same model vary significantly among sections.}
\label{table:evaluation_ipc}
\end{table*}

%% file: tabs/abstract_evaluation.tex
\begin{table}[t!]
\footnotesize
\centering
\begin{tabular}{lcccccccc}
\toprule

 \textbf{Model} & \textbf{BLEU} & \textbf{R-1} & \textbf{R-L} & \textbf{BERTScore}  \\
\midrule
Patent-GPT-J &  7.71 & 22.46 & 18.07 & 79.23 \\
GPT-4 &  15.11 & 46.78 & 30.95 & 87.61 \\
Llama-3 & 24.09 & 50.93 & 40.08 & 88.65 \\
Llama-3-FT & 48.48 & 67.90 & 59.66 & 92.13 \\
Llama-3-MFT & 48.49 & 67.69 & 59.65 & 92.13 \\
\bottomrule
\end{tabular}
\caption{Evaluation results of abstract}
\label{table:evaluation_abstract}
\end{table}

%% file: tabs/annotation_criteria.tex
\begin{table*}[p]
\centering
\footnotesize
\begin{tabular}{|>{\raggedright\arraybackslash}p{3cm}|>{\raggedright\arraybackslash}p{12cm}|}
\toprule
\textbf{Criteria} & \textbf{Rating Description} \\
\midrule

\textbf{Completeness of Essential Features} &
\begin{itemize}
    \item \textbf{0-2:} Most essential features are missing or poorly described.
    \item \textbf{3-4:} Some essential features are present but significant gaps remain.
    \item \textbf{5-6:} Majority of essential features are covered but with minor omissions.
    \item \textbf{7-8:} Almost all essential features are well described with very few gaps.
    \item \textbf{9-10:} All essential features are thoroughly and comprehensively covered.
\end{itemize} \\
\midrule

\textbf{Conceptual Clarity} &
\begin{itemize}
    \item \textbf{0-2:} Claims are very unclear and ambiguous.
    \item \textbf{3-4:} Claims have significant clarity issues, making them difficult to understand.
    \item \textbf{5-6:} Claims are mostly clear but contain some ambiguous language.
    \item \textbf{7-8:} Claims are clear with minimal ambiguity.
    \item \textbf{9-10:} Claims are exceptionally clear and completely unambiguous.
\end{itemize} \\
\midrule

\textbf{Consistency in Terminology} &
\begin{itemize}
    \item \textbf{0-2:} Terminology is highly inconsistent.
    \item \textbf{3-4:} Significant inconsistencies in terminology.
    \item \textbf{5-6:} Some inconsistencies in terminology but mostly uniform.
    \item \textbf{7-8:} Terminology is largely consistent with minor inconsistencies.
    \item \textbf{9-10:} Terminology is completely consistent throughout.
\end{itemize} \\
\midrule

\textbf{Technical Correctness of Feature Linkages} &
\begin{itemize}
    \item \textbf{0-2:} Features are poorly linked with many inaccuracies.
    \item \textbf{3-4:} Significant issues with the linkages of features.
    \item \textbf{5-6:} Mostly accurate linkages with some incorrect connections.
    \item \textbf{7-8:} Accurate linkages with minor inaccuracies.
    \item \textbf{9-10:} Features are accurately and correctly linked throughout.
\end{itemize} \\
\midrule

\textbf{Overall Quality} &
\begin{itemize}
    \item Calculated by: $(completeness*4 + clarity*2 + consistency*2 + correctness*3) \div 11$
    \item \textbf{0-2:} Very poor overall quality, fails to meet most criteria.
    \item \textbf{3-4:} Low overall quality with significant issues across multiple criteria.
    \item \textbf{5-6:} Average overall quality, meets criteria at a basic level.
    \item \textbf{7-8:} High overall quality with minor issues.
    \item \textbf{9-10:} Excellent overall quality, meets or exceeds all criteria.
\end{itemize} \\
\bottomrule

\end{tabular}
\caption{Rating criteria for human annotation}
\label{tab:humanrating}
\end{table*}